\documentclass[11pt,a4paper]{article}
\usepackage[a4paper,margin=2.2cm]{geometry}

\usepackage[utf8]{inputenc}
\usepackage[T1]{fontenc}

\usepackage{mathptmx} 

\usepackage{setspace}
\usepackage{caption}
\usepackage{graphicx}
\usepackage{amsmath}
\usepackage{amssymb}
\usepackage{booktabs}

\usepackage{todonotes}
\newlength{\extralength}
\usepackage{changepage}
\usepackage{tabularx}
\usepackage{multirow}
\usepackage{threeparttable}

\usepackage{authblk}
\usepackage{orcidlink}

\title{Real-Time Structural Detection for Indoor Navigation from 3D LiDAR Using Bird's-Eye-View Images}

\author{Guanliang Li\orcidlink{0009-0002-9736-8620}}
\author{Pedro Espinosa-Angulo\orcidlink{0009-0008-2079-8650}}
\author{David Perez-Saura\orcidlink{0000-0003-2571-3165}}
\author{Santiago Tapia-Fernández\orcidlink{0000-0002-1418-9276}}
\affil{Universidad Politécnica de Madrid \\ 
\texttt{guanliang.li@alumnos.upm.es}, \texttt{pedro.espinosa@upm.es}\\
\texttt{david.perez.saura@upm.es}, \texttt{santiago.tapia@upm.es}}

\begin{document}

\maketitle

\abstract{Efficient structural perception is essential for mapping and autonomous navigation on resource-constrained robots. Existing 3D methods are computationally  prohibitive, while traditional 2D geometric approaches lack robustness. This paper presents a lightweight, real-time framework that  projects 3D LiDAR data into 2D Bird’s-Eye-View (BEV) images to enable efficient detection of structural elements relevant to mapping and navigation. Within this representation, we systematically evaluate several feature extraction strategies, including classical geometric techniques (Hough Transform, RANSAC, and LSD) and a deep learning detector based on YOLO-OBB. The resulting detections are integrated through a spatiotemporal fusion module that improves stability and robustness across consecutive frames. Experiments conducted on a standard mobile robotic platform highlight clear performance trade-offs. Classical methods such as Hough and LSD provide fast responses but exhibit strong sensitivity to noise, with LSD producing excessive segment fragmentation that leads to system congestion. RANSAC offers improved robustness but fails to meet real-time constraints. In contrast, the YOLO-OBB-based approach achieves the best balance between robustness and computational efficiency, maintaining an end-to-end latency (satisfying 10 Hz operation) while effectively filtering cluttered observations in a low-power single-board computer (SBC) without using GPU acceleration. The main contribution of this work is a computationally efficient BEV-based perception pipeline enabling reliable real-time structural detection from 3D LiDAR on resource-constrained robotic platforms that cannot rely on GPU-intensive processing. The source code and pre-trained models are publicly available.}

\section{Introduction}
\label{sec:introduction}

With the rapid advancement of robotics, Indoor Mobile Robots (IMR) have demonstrated substantial potential in diverse applications, including industrial logistics, household service, and commercial guidance. To achieve autonomous operation in these complex settings, robots must rely on Simultaneous Localization and Mapping (SLAM) to perceive their surroundings in real-time\cite{deepa2024design, zhang2022research, li2024robot}. While traditional SLAM systems effectively construct occupancy grid maps for collision avoidance, advanced autonomous tasks require a higher level of environmental understanding, specifically semantic structural mapping. Identifying structural elements such as walls and rooms is critical for semantic navigation, logical path planning, and human-robot interaction. With the advancement of sensor technology, 3D Light Detection and Ranging (LiDAR) has emerged as the dominant sensor for indoor perception due to its superior depth precision and robustness against illumination changes compared to cameras\cite{NIE2026119639, wu2025indoor, tee2021lidar}.

However, processing high-fidelity 3D LiDAR data presents a significant challenge for mobile robots equipped with resource-constrained onboard computers. Existing approaches typically fall into two extremes. On the one hand, 3D semantic segmentation networks offer high accuracy but incur massive computational overheads, often requiring high-end GPUs that are impractical for battery-powered embedded systems due to power and weight constraints. On the other hand, traditional geometric methods operating on 2D slices are computationally efficient but lack robustness. In cluttered indoor environments filled with furniture and dynamic objects, these methods often fail to distinguish between structural walls and transient obstacles, leading to noisy and fragmented maps. Consequently, there is a lack of a perception solution that can achieve 3D-level robustness while maintaining 2D-level computational efficiency.

To address this gap, this paper proposes a lightweight real-time room detection module designed specifically for the robotic systems onboard. Our core strategy utilizes dimensionality reduction: we project the rich 3D point cloud into a 2D Bird's-Eye-View (BEV) representation, effectively filtering out ceiling and ground noise while compressing the data volume. Based on this framework, we introduce a modular pipeline to extract structural features. We implement and systematically evaluate a spectrum of lightweight methods, ranging from traditional geometric fitting algorithms---including Random Sample Consensus (RANSAC) and the Hough Transform---to modern computer vision and deep learning techniques, specifically the Line Segment Detector (LSD) and YOLO. By directly comparing the performance of these diverse approaches, we identify the optimal solution that best balances geometric precision with computational efficiency for embedded applications.

The main contributions of this paper are summarized as follows:

\begin{enumerate}
    \item \textbf{Design of a Lightweight Perception Pipeline:} We propose an end-to-end framework that transforms raw 3D LiDAR data into a semantic vector map. Using BEV projection and modular design (ROS2), the system achieves real-time performance on resource-constrained hardware.
    \item \textbf{Comparative Analysis of Extraction Paradigms:} We conduct a systematic evaluation of multiple feature extraction strategies, including RANSAC, the Hough Transform, LSD, and YOLO (You Only Look Once). The analysis highlights the trade-offs between inference speed, noise immunity, and geometric precision.
    \item \textbf{Robust Spatiotemporal Fusion:} We implement a lightweight post-processing module that fuses single-frame detections using temporal data association and lifecycle management. This ensures the generation of a globally consistent and clean room structure map, effectively filtering out dynamic clutter.
\end{enumerate}

The remainder of this paper is organized as follows: Section 2 reviews related work in LiDAR-based mapping. Section 3 outlines the high-level system architecture. Section 4 details the specific algorithms for data processing and feature extraction. Section 5 presents the experimental validation and performance analysis in different scenarios. Finally, Section 6 concludes the study.
\section{Related work}
\label{sec:related_work}

This section provides a comprehensive review of mainstream methodologies for feature extraction and semantic understanding based on the 3D point cloud. We begin by examining deep learning approaches, specifically analyzing the trade-off between detection accuracy and computational cost. Subsequently, we explore traditional geometric feature extraction algorithms, highlighting their limitations when applied to complex, unstructured environments. Finally, the discussion focuses on projection-based computer vision techniques. We offer a comparative analysis of three specific paradigms—line detection, semantic segmentation, and object detection—evaluating their applicability and performance within real-time airborne systems.

\subsection{3D Point Cloud Semantic Understanding}
\label{sub:3D_point_cloud_semantic_understanding}

The advent of deep learning has significantly enhanced the capability of machines to semantically understand three-dimensional space. PointNet \cite{qi2017pointnet} represents a pioneering approach that directly processes unstructured point clouds. By utilizing shared Multi-Layer Perceptrons (MLPs) to extract point-wise features independently and employing max pooling to aggregate global information, it achieves end-to-end classification and segmentation. Its successor, PointNet++ \cite{qi2017pointnet++}, introduced a hierarchical feature learning mechanism that aggregates local neighborhood features at multiple scales, thereby significantly enhancing the recognition of fine-grained patterns. Despite their superior accuracy, these methods rely heavily on k-Nearest-Neighbor (k-NN) search and Ball Query operations. These operations typically result in super-linear growth in computational complexity, which hinders real-time execution on low-power airborne platforms \cite{sarker2024comprehensive}. 

Conversely, Voxel-based Convolutional Neural Networks (CNNs) attempt to transform irregular point clouds into regular grids to facilitate the application of efficient 3D convolutions. VoxNet \cite{maturana2015voxnet} was among the first to voxelize point clouds before feeding them into a 3D CNN for object recognition. OctNet \cite{riegler2017octnet} further optimized storage efficiency in sparse spaces by utilizing octree structures. However, voxelization methods face an inherent trade-off between resolution and memory consumption: low resolution leads to information loss, while high resolution results in prohibitive memory usage. Although recent work such as PVCNN \cite{liu2019point} attempts to combine the advantages of point-based and voxel-based representations to reduce computational overhead, deploying these GPU-dependent 3D networks on resource-constrained embedded platforms remains a significant challenge.

In summary, while processing 3D data directly preserves rich spatial information, the associated high computational demand renders it unsuitable as a primary solution for lightweight airborne systems.

\subsection{Geometric Method for Features Extraction}
\label{sub:geometric_method_for_features_extraction}

Traditional geometric methods operate on the assumption that environmental structures, such as walls, adhere to specific mathematical models (e.g., planes or lines). These methods typically estimate parameters from noisy data through statistical iteration. Paradigmatic algorithms in this category include RANSAC and the Hough Transform.

RANSAC \cite{fischler1981random} is a robust parameter estimation technique that constructs hypothetical models through random sampling and validates the consensus set (inliers) to reject outliers. Due to its inherent robustness against noise, RANSAC is widely employed for plane fitting in point clouds. Although Schnabel et al. \cite{schnabel2007efficient} and Li et al. \cite{li2017improved} have optimized its efficiency by introducing normal vector constraints and grid preprocessing, the performance of RANSAC remains highly dependent on the number of iterations. In cluttered indoor environments, a large number of unstructured objects (e.g., furniture and equipment) constitute a significant proportion of outliers. This forces the algorithm to perform excessive invalid iterations, resulting in uncontrollable computational latency. Furthermore, RANSAC lacks semantic understanding capabilities and is prone to misclassifying linearly aligned obstacles as structural walls.

The Hough Transform \cite{hough1962method}, conversely, converts the detection problem into a peak search within a parameter space. It is particularly effective in detecting boundaries that are fragmented due to occlusion. However, its core limitation lies in the quantization of the parameter space. To ensure high precision in wall localization, an extremely fine parameter grid is required. This leads to an exponential increase in memory consumption, making it difficult to meet the real-time requirements of embedded systems \cite{borrmann20113d}.

Consequently, while geometric methods are efficient in simple scenarios, they face dual challenges with regard to computational efficiency and semantic discriminability in complex environments, which fails to provide high-level semantic information.

\subsection{Image Based Feature Extraction Paradigms}
\label{sub:feature_extraction_paradigms}

To strike a balance between computational efficiency and perceptual capability, projecting 3D point clouds into 2D representations (such as BEV or Range Images) and applying mature computer vision techniques have become a prevailing trend. Based on the output format, these approaches can be categorized into three main paradigms: traditional line detection, semantic segmentation, and object detection.

\subsubsection{Traditional Line Detection}
\label{subsub:traditional_line_detection}

The LSD \cite{von2008lsd} is a linear feature extraction algorithm based on image gradients, characterized by its parameter-free nature and high computational efficiency. Several studies have attempted to apply LSD to point cloud projections \cite{du2023fast}, or to combine it with deep learning to optimize gradient fields (e.g., DeepLSD \cite{pautrat2023deeplsd}). However, in complex indoor environments, BEV images often contain significant high-frequency noise. Lacking semantic awareness, LSD cannot distinguish between "wall edges" and "clutter edges" (e.g., furniture). This limitation results in detection outputs populated with fragmented false positives, which severely degrades the quality of the resulting map.

\subsubsection{Semantic Segmentation Networks}
\label{subsub:semantic_segmentation_networks}

With the advancement of Convolutional Neural Networks (CNNs), pixel-wise classification methods have become a popular choice for extracting environmental features. Prominent examples include U-Net applied to BEV maps \cite{gao2024damage}, or SqueezeSeg \cite{wu2018squeezeseg} and RangeNet++ \cite{milioto2019rangenet++} applied to Range Images. These models effectively capture local context to achieve end-to-end semantic segmentation. However, the output of segmentation models inherently consists of unstructured pixel masks. To obtain the geometric parameters required for vector mapping (such as wall segments), the system necessitates complex post-processing steps, such as skeletonization or contour fitting. This "segmentation-plus-fitting" pipeline not only introduces computational redundancy but is also prone to discretization errors, making it difficult to meet the stringent requirements of high-precision vector mapping.

\subsubsection{Object Detection Networks}
\label{subsub:object_detection_networks}

To overcome the aforementioned limitations, object detection offers a more direct solution. The YOLO series \cite{hussain2024yolov5}, renowned for its efficient architecture, has become one of the best choices for real-time vision tasks. Unlike segmentation networks, YOLO is an end-to-end detector. Specifically, the Oriented Bounding Box (OBB) version is capable of directly regressing the geometric parameters of walls (center, dimensions, and angle) from point cloud BEV maps \cite{khanam2024yolov11}. This output format provides "out-of-the-box" utility, eliminating the need for complex post-processing. Furthermore, the trained neural network has inherent noise suppression capabilities, enabling robust extraction of structural features from cluttered backgrounds. Although the computational load is slightly higher than that of purely geometric methods, it is fully capable of meeting real-time requirements.

Above all, traditional LSD lacks robustness against noise, while segmentation-based methods suffer from post-processing bottlenecks. In contrast, the YOLO-OBB detection scheme achieves an optimal trade-off between data structure consistency, noise immunity, and inference speed. Consequently, it has been designated as the core feature extraction module for this study.

\section{Room Structure Elements Detection Module Design}
\label{sec:detection_module}

\begin{figure}[htbp]
    \centering
    \includegraphics[width=0.8\textwidth]{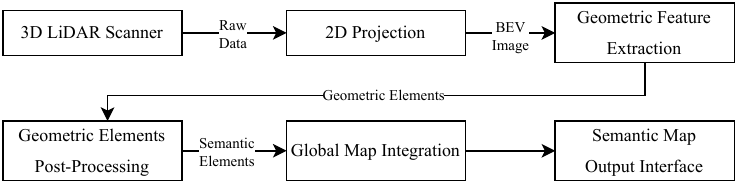}
    \caption[Wall and Room Recognition Module Architecture]{\textbf{Wall and Room Recognition Module Architecture.} The pipeline details the modular process of transforming raw 3D LiDAR data into semantic structural elements via 2D projection and geometric feature extraction.}
    \label{fig:Wall_and_Room_Recognition_Module_Architecture}
\end{figure}

The proposed detection model is designed to seek a good balance between functionality, robustness requirements, and limited computational resources; the target is to provide a detection pipeline that can be executed in a mini-PCs or in Single Board Computers (SBCs) that can be mounted on robots. The diagram in Figure~\ref{fig:Wall_and_Room_Recognition_Module_Architecture} illustrates the module pipeline.

The main idea for this pipeline is a mapping strategy based on dimension-reduction as the first step and then extraction of interference-resistant features. Rather than operating directly on 3D point clouds, this strategy processes raw range data from LiDAR sensors to produce the point cloud in cylindrical coordinates. Since the azimuthal angles are fixed for a given sensor and the height is not absolutely necessary for a planar map, only one coordinate, the radius, is relevant to the problem. In addition to this transformation, the point cloud is represented as an image in which the LiDAR sensor is located at the center of the image and each point in the cloud is transformed into a black corresponding pixel whose indexes are computed using the radius and a scale factor between pixels and length. Thus, the point cloud is compressed into a 2D BEV, significantly reducing the data size and enabling our pipeline to detect structural objects using a computer vision approach.

After this preprocessing step, the system uses the subsequent feature extraction algorithms on the images to identify geometric features and utilizes a map construction module to further eliminate errors, ultimately constructing a clear and accurate floor map. Since there are plenty of computer vision algorithms to extract these features and their performance in terms of time-execution and outcome quality cannot easily be predicted, the pipeline is designed to use any of them. To do so, the system architecture is based on ROS2, each algorithm has been rosified into a node that can be easily replaced by another alternative, thus providing flexibility and allowing a fair comparison environment for assessing the algorithms.  

The geometric feature extraction module is dedicated to structural environmental perception and topology extraction, namely, it looks for walls, columns, corners, and/or other structural objects surrounding the robot. Actually, this module consists of a set of nodes that implement algorithms from classical computer vision and neural networks models that process the BEV images and extract the location for the identified objects. 

After identifying elements in the BEV image, the elements found are preprocessed to produce a data structure of semantic elements, that is, a data structure that contains mainly walls, columns, and other structural elements.  The important thing about this module is that the element positions are not computed related to the sensor position; instead the positions of the elements are computed related to each other. Using this approach produces a graph as the data structure that contains the elements and their relative positions. This graph and its values could not change very much over time since the structural elements do not move; in fact, changes will be a consequence of sensor noise and algorithm inaccuracy in extracting the geometric information. So improving the precision could be achieved just by very simple filtering methods. 

The global map integration is a node for integrating the information of successively BEV images and their analysis; the mission of this module is to build a map to track the structural element positions as time passes. In doing so, it will be able to produce a stable identification of structural elements. The elements will be identified by numbers and categories; that is, the same real element will be identified using the same number and category over time, even if the robot moves in its environment. 

Finally, the last module, an output interface for other mapping software, prepares all the computed data to be used in a large system. This pipeline could be used both as a stand-alone basic SLAM module or as a detection subsystem in a full and more complex Semantic SLAM system. As a stand-alone software, the information from the Global Map Integration could be used to build a map and estimate the robot position, while this module could produce a list of identified objects and their relative position to the sensor that can be used in a SSLAM system using a graph-based optimization. 

A full description of each stage is given in the next section. 

\section{Feature Extraction and Mapping Methodology}
\label{sec:feature-extraction}

Building upon the modular system architecture proposed in Section 3, this section details the implementation of the perception pipeline designed to maximize computational efficiency on resource-constrained onboard platforms. Consistent with the analysis in \ref{sub:3D_point_cloud_semantic_understanding}, 3D point cloud semantic understanding networks are computationally intensive, therefore, excluded from this study. Instead, we focus on lightweight strategies that operate on projected 2D representations to achieve real-time performance at a frequency of more than 10Hz that is more than the common frequency of LiDAR.

\begin{figure}[htbp]
    \centering
    \includegraphics[width=\textwidth]{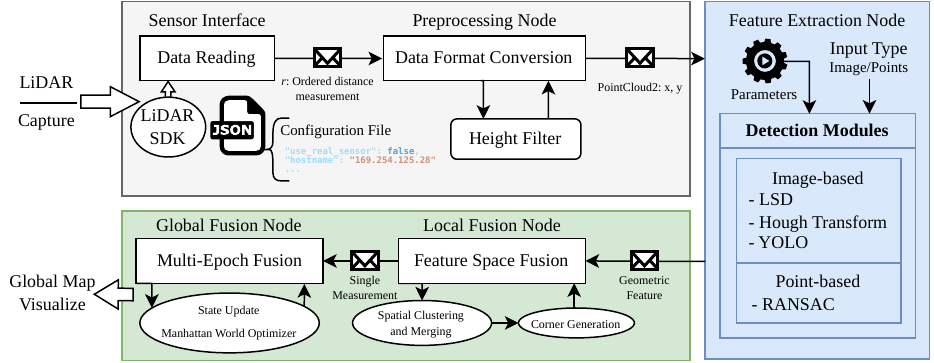}
    \caption[System Architecture Diagram]{\textbf{System Architecture Diagram.} The system comprises five ROS2 nodes, categorized into three functional modules based on the data processing pipeline: Data Interface, Feature Detection, and Feature Fusion.}
    \label{fig:Node_Schema}
\end{figure}

The methodology is structured following the system's data flow as shown in Figure~\ref{fig:Node_Schema}. First, we introduce the dimensionality reduction technique that transforms raw 3D LiDAR data into 2D feature spaces. Subsequently, we select a set of lightweight feature extraction algorithms spanning from traditional geometric fitting to modern learning-based detection and detail their specific implementations. As discussed in Section \ref{sec:related_work}, specific implementations include RANSAC and the Hough Transform, the LSD, and YOLOv8n-obb. These algorithms were chosen to represent the paradigms of statistical parameter estimation, image gradient analysis, and deep object detection, respectively. Finally, we describe the data post-processing framework designed to filter noise and integrate discrete detection results into a globally consistent semantic map.

\subsection{Point Cloud Flattening}
\label{sub:pointcloud-flatten}

For the specific detection target of vertical walls, leveraging their geometric consistency in the vertical direction, this section proposes a dimensionality reduction preprocessing algorithm based on height compression. This method projects the three-dimensional spatial information along the \( z \)-axis onto a two-dimensional plane, thus transforming the complex problem of 3D point cloud segmentation into a feature extraction task within a 2D plane. The strategy fully utilizes the measurement error characteristics of LiDAR--specifically, when multiple laser beams strike the same vertical wall surface, the projected 2D point cloud exhibits a band-shaped distribution with a certain thickness due to measurement noise. This statistical feature greatly facilitates subsequent point cloud segmentation and edge identification. Furthermore, through dimensionality reduction, the method significantly decreases the algorithm's time complexity and hardware computational requirements.

Specifically, the algorithm consists of the following three core steps:

\begin{enumerate}
    \item \textbf{3D Cartesian Transformation:} \\
    Based on the inherent horizontal azimuth and vertical elevation parameters of the LiDAR's array, the raw distance data \( R \) output by the sensor is mapped from the spherical coordinate system to 3D Cartesian coordinates \( (x, y, z) \), thus restoring the true spatial structure of the environment.

    \item \textbf{Height Threshold Filtering:} \\
    In the 3D LiDAR scanning pattern, the ground and ceiling typically appear as high-density concentric ring-shaped point clouds, which pose significant interference to the detection of lateral walls. Using height information on the \( z \)-axis obtained in Step 1, the algorithm defines a Region of Interest (ROI) and directly removes point cloud data whose \( z \) values are outside a set range allowing to filter ground and ceiling points, retaining only data representing potential vertical structures.

    \item \textbf{2D Projection Generation and Storage:} \\
    The \( (x, y) \) coordinates of the resulting point cloud are extracted and reorganized into a 2D point set, which is then stored in memory. This standardized 2D data format exhibits strong compatibility and can serve directly as the input foundation for subsequent modules.
\end{enumerate}

Through the above process, this algorithm effectively avoids the high computational cost associated with 3D spatial calculations. While substantially reducing data dimensionality, it fully preserves the key geometric features (vertical wall contours) in the environment, laying the foundation for real-time detection.

\subsection{Feature Extraction Algorithms}
\label{sub:feature-extraction-algorithms}

The processing of low-dimensional point clouds depends on the nature of the algorithm used. In this study, we categorize the methods into two main approaches: (1) direct point-based processing and (2) image-based processing via BEV projection.

The first approach is utilized by RANSAC, which operates directly on the discrete \((x,y)\) coordinates of the point cloud to fit mathematical models without the need for rasterization.

The second approach involves converting point clouds into a 2D image. This step is mandatory for YOLO and LSD, as they are inherently computer vision algorithms designed to extract features from pixel grids. Additionally, while the Hough Transform can theoretically operate on discrete points, we classify it under the image-based approach in this implementation. This is to leverage the highly optimized Probabilistic Hough Transform provided by the OpenCV library, which requires a binary image input.

The generated image serves as an occupancy grid, forming a binary image combined with the detected point cloud. Given that the LiDAR resolution is approximately 2-3 cm, each square pixel in the grid is defined to represent this spatial range. However, using an occupancy grid can lead to high computational costs if not properly managed, as the total number of cells grows quadratically with linear increases in the dimensions of the map.

Consequently, to balance computational resource consumption with operational requirements, the maximum scene range is set to approximately 40 meters, which corresponds to an image resolution of 4096 × 4096 pixels. The following sections introduce various detection methods.

\subsubsection{DBSCAN-aided RANSAC}
\label{subsub:ransac}

To address the computational inefficiency and sensitivity to outliers  inherent in global RANSAC applications, this study implements a hybrid two-stage pipeline combining Density-Based Spatial Clustering of Applications with Noise (DBSCAN) and RANSAC. Instead of performing a blind search across the entire point cloud,  this strategy decomposes the feature extraction task into coarse-grained clustering followed by fine-grained fitting.

First, DBSCAN is utilized to perform density-based clustering on the 2D  projected point cloud. This step effectively separates high-density  structural regions (potential walls and obstacles) from sparse  environmental noise, grouping spatially adjacent points into independent clusters. By answering the question of "which points belong to the same object," DBSCAN acts as a spatial filter that reorganizes the  unstructured global data into discrete local subsets. In practical operation, it is necessary to set the neighborhood radius and the minimum neighbor number threshold based on the LiDAR resolution and scene range to achieve correct clustering and improve the RANSAC running speed. Therefore, this method has the disadvantage that it is not able to automatically achieve full scene coverage.

Subsequently, the RANSAC algorithm with the minimum number of sampling points and residual threshold set is applied to the identified clusters. Within these constrained subsets, RANSAC fits linear models to  extract precise wall contours and outputs the optimal line segment  parameters. This \textit{divide-and-conquer} approach significantly reduces the search space for geometric fitting, thereby lowering the required  number of iterations. Furthermore, since DBSCAN pre-filters discrete  noise, the input quality for RANSAC is improved, enhancing both the  stability of the model and the accuracy of the final wall  representation.

\subsubsection{Hough Transform}
\label{subsub:hough-transform}

For image-based geometric extraction, this study employs the Probabilistic Hough Transform (PHT) to identify linear structural elements. Unlike the standard Hough Transform, which maps every edge pixel to the parameter space, the probabilistic variant minimizes computational load by analyzing a random subset of points sufficient to detect line features. This efficiency is critical for the real-time constraints of the system. The PHT algorithm in the OpenCV library is directly adopted with the feature space unit and the minimum line segment length set. Finally, the detected line segments defined in the pixel coordinate system are inversely transformed into the LiDAR coordinate system to populate the spatial feature set.

Regarding its use in this specific context, it is simpler than in other scenarios. While the target image is typically subjected to preprocessing, such as the Canny edge detector, this step is unnecessary in the present case because the images are binary.

\subsubsection{LSD}
\label{subsub:lsd}

As a comparative alternative to the voting-based Hough Transform, LSD is implemented to extract features based on image gradients. This algorithm is characterized by its linear-time execution and parameter-free nature, which avoids the manual tuning of thresholds required by other geometric methods. In this study, the standard LSD implementation from the OpenCV library is utilized, configured with standard refinement to improve the precision of the endpoint. Similarly to the PHT pipeline, the detector processes the BEV image directly without the need for prior edge extraction, generating a list of segments in pixel coordinates. These are subsequently mapped back to the robot's metric reference frame via the intrinsic resolution parameters, producing a collection of geometric constraints for the mapping system.

Note that both Hough Transform and LSD identify wall edges, and the detection results need to be further fused and skeletonized to achieve accurate wall position recognition.

\subsubsection{YOLO}

YOLO is a widely recognized neural network architecture in the field of deep learning and computer vision. It is primarily used for object detection within an image. Since its introduction, numerous architectures have been proposed to improve on previous versions, continuously advancing its capabilities.

The use of YOLO in the workplace is a consequence of its advantages as a noise-resistant method with the ability to detect objects of multiple classes. The version used is YOLOv8-obb because, in order to obtain the coordinates of the structural elements of the scene with the highest possible precision, it is essential to have oriented bounding boxes to correctly capture elements regardless of their spatial position.

The restriction on the map dimensions ($4096 \times 4096$ pixels) is based on the intended operational environment. From a modeling perspective, given that the standard input size for the trained YOLO models is $1024 \times 1024$ pixels, the selected map resolution ensures that the image downsampling factor does not exceed four. Resizing beyond this threshold would result in an excessive loss of fine-grained detail, which is considered detrimental to detection precision.

Nevertheless, the advantages associated with YOLO—similar to those of other neural models—cannot be achieved without a sufficiently extensive and diverse dataset that allows training in various situations, such as occlusions, noise, and different rotations and positions of objects. This represents a disadvantage compared to alternatives such as the Hough transform or LSD, since, to our best knowledge, there are no annotated datasets of BEV (Bird’s Eye View) images obtained with the type of sensor we have available. Therefore, the creation of the dataset was part of the work carried out for this article.

\begin{figure}[htbp]
    \centering
    \includegraphics[width=0.8\textwidth]{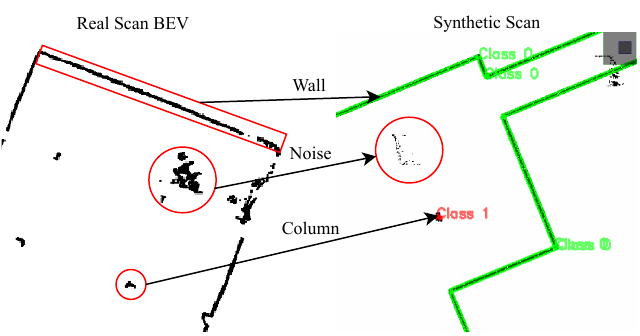}
    \caption[Synthetic Data Generation Pipeline]{\textbf{Qualitative Comparison between a Real LiDAR BEV Scan and a Generated Synthetic Training Sample.} The correspondence arrows illustrate the high fidelity of the proposed generation pipeline in mimicking real-world sensor characteristics: (Top) Continuous wall structures are automatically annotated with green OBBs (Class 0); (Middle) Real-world noise patterns are injected to simulate sensor artifacts; (Bottom) Discrete obstacles like columns are instantiated with red OBBs (Class 1). This visual similarity ensures a minimal domain gap for model training.}
    \label{fig:Synthetic_Data_Generation_Pipeline}
\end{figure}

The dataset used for training consists of synthetic images generated by a digital program. The rationale behind using synthetic image generators is their ability to produce a vast variety of scenes that would be impractical to collect through real-world data alone. Beyond the inherent limitations of real-world scene diversity, real images also require manual annotation, which significantly increases the cost of dataset creation. Conversely, synthetic image generation enables fully automated annotation, allowing entire datasets to be generated in mere minutes. A comparison between synthetic and real scans is shown in Figure~\ref{fig:Synthetic_Data_Generation_Pipeline}, where the incorporation of noise simulation further enhances the model's robustness and bridges the gap between reality and simulation.

In most cases, synthetic image generation is complex and challenging, but it is relatively straightforward in this context because indoor BEV images are highly predictable. These images primarily consist of well-defined rooms bounded by walls that typically form 90° intersections, columns, and obstacles within them.
To enhance the robustness of wall detection, real-world noise patches—obtained from actual scenes—have been incorporated into synthetic images, either applied randomly across the entire image or overlaid specifically on walls. Training results obtained using YOLO Nano—due to the need for high inference speed on limited hardware—demonstrate strong performance, with performance degradation being minimal when transitioning to real-world scenarios. This indicates that the proposed data generation approach maintains a reasonable resemblance to real-world conditions, thereby greatly reducing the necessity for domain adaptation techniques when deploying the trained model.

\subsection{Data Post-processing}
\label{sub:data-postprocessing}

\begin{figure}[htbp]
    \centering
    \includegraphics[width=0.8\textwidth]{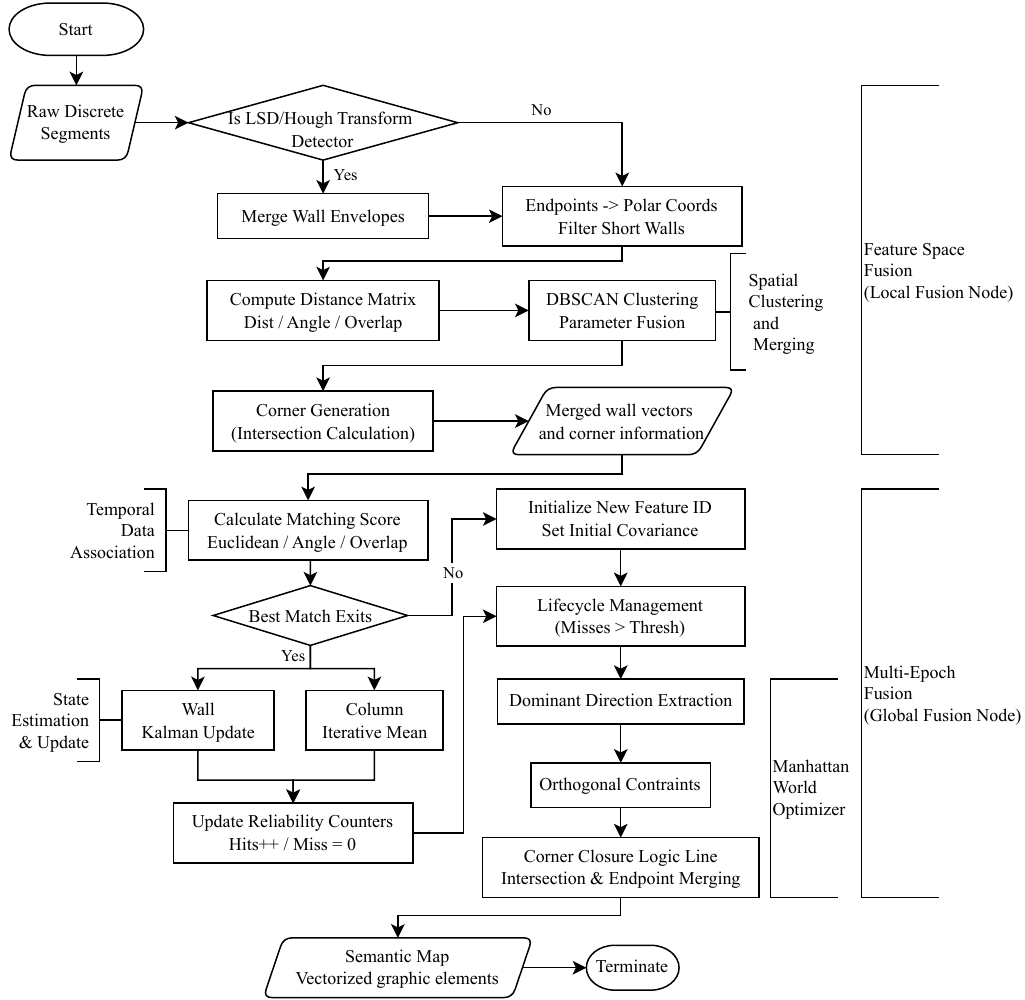}
    \caption[Flowchart of the Data Post-processing and Fusion Algorithm]{\textbf{Flowchart of the Data Post-processing and Fusion Algorithm.} The process is divided into two main stages: the Local Fusion Node for feature extraction and spatial merging, and the Global Fusion Node for multi-epoch data fusion and noise filtering.}
    \label{fig:Flowchart_of_the_Data_Post-processing_and_Fusion_Algorithm}
\end{figure}

The output of the feature extraction modules described above typically consists of discrete line segments. However, because of environmental occlusions and inherent measurement noise, these raw detections often suffer from fragmentation and instability. Consequently, a post-processing module is essential to reconstruct complete wall structures from fragmented segments and to enhance localization accuracy through data fusion. The primary objective of this module is to unify fragmented wall segments and integrate detection results across multiple temporal epochs (frames) to generate a high-precision global map. The overall workflow is illustrated in Figure~\ref{fig:Flowchart_of_the_Data_Post-processing_and_Fusion_Algorithm}, which organizes the processing pipeline into two main nodes: Feature Space Fusion and Multi-Epoch Fusion. 

\subsubsection{Feature Space Fusion}

The primary objective of the Local Fusion Node is to unify fragmented segments detected in the current BEV frame into a concise set of geometric primitives.

\begin{enumerate}
    \item Detector-Specific Preprocessing:
    The pipeline first handles different detector characteristics. Since LSD and Hough Transform tend to generate wall envelopes which are composed of double edges, a \textit{Merge Wall Envelopes} step is applied to consolidate these structures into a single representative line, as illustrated in Figure~\ref{fig:merge_envelopes}.

    \begin{figure}[htbp]
        \centering
        \includegraphics[width=0.8\textwidth]{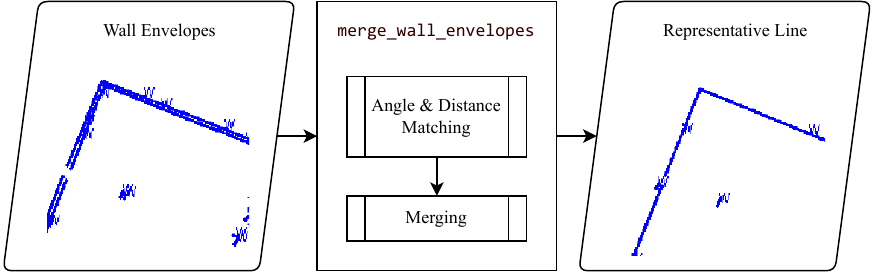}
        \caption[Merge Wall Envelopes for LSD and Hough Transform]{\textbf{Merge Wall Envelopes for LSD and Hough Transform.} Illustration of the merge wall envelopes step, consolidating double-edge structures into a single representative line.}
        \label{fig:merge_envelopes}
    \end{figure}

    \item Polar Parameterization and Filtering:
    Each detected line segment, originally represented by its endpoints $(x_1, y_1, x_2, y_2)$, is transformed into a polar representation $(\rho, \alpha, d_1, d_2)$. Here, $\rho$ is the perpendicular distance from the origin to the infinite line containing the segment, and $\alpha$ is the angle of the line's normal vector (i.e., the direction perpendicular to the line). Parameters $d_1, d_2$ are the signed distances from the foot of the perpendicular point to the two endpoints, measured along the line direction. This representation avoids the singularity of vertical lines and naturally decouples the line's orientation from the extent of the segment. Segments whose length $|d_2 - d_1|$ falls below a predefined threshold $\tau_{len}$ are discarded as noise, retaining only those that are long enough to represent meaningful wall structures.
    
    \item Spatial Clustering with a Custom Distance Metric:
    To group fragmented detections belonging to the same physical wall, we employ DBSCAN with a precomputed distance matrix. The distance between two lines segments \textit{i} and \textit{j} is defined as the maximum of three normalized components:
    
    \begin{equation}
    D_{ij} = \max\left( \frac{\Delta d}{\tau_d}, \frac{\Delta\theta}{\tau_\theta}, \frac{\Delta o}{\tau_o} \right),
    \end{equation}
    
    where:
    \begin{itemize}
    \item $\Delta d$ is the Euclidean distance between the two segments, calculated as the minimum of the four point-to-line distances from the end points of one segment to the infinite line of the other. This captures spatial proximity.
    \item $\Delta\theta$ is the angular difference between the segment orientations, accounting for wrap-around by taking $\Delta\theta = \min\bigl(|\theta_i - \theta_j|,\; \pi - |\theta_i - \theta_j|\bigr)$.
    \item $\Delta o$ quantifies the overlap between the two segments along their common direction. It is calculated as the distance between their projected intervals in the line direction. This prevents discontinuous walls from being incorrectly merged.
    \end{itemize}
    The above three elements are further explained in Figure~\ref{fig:clustering_metrics}.

    \begin{figure}[htbp]
        \centering
        \includegraphics[width=\textwidth]{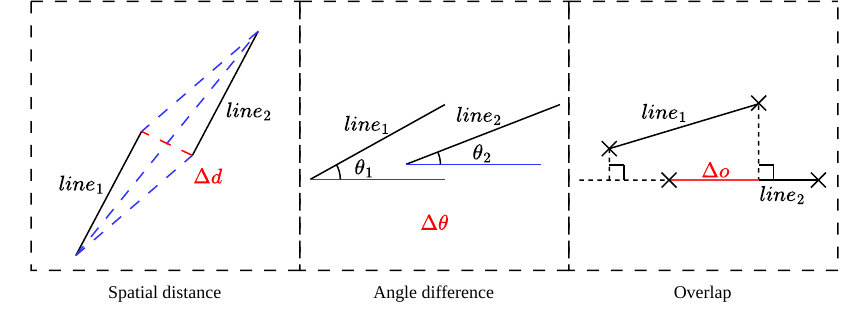}
        \caption[Illustration of the Clustering Metrics]{\textbf{Illustration of the CLustering Metrics.} Depiction of the three components used in the custom distance metric for DBSCAN: spatial distance $\Delta d$, angular difference $\Delta\theta$, and segment overlap $\Delta o$.}
        \label{fig:clustering_metrics}
    \end{figure}
    
    The thresholds $\tau_d$, $\tau_\theta$, and $\tau_o$, set based on the characteristics of the sensor and the layout of the environment, are used to normalize the errors. By taking the maximum of the three terms, the metric ensures that only segments that are simultaneously close in position, similar in orientation, and with sufficient overlap are considered neighbors.
    
     After obtaining the distance matrix, DBSCAN groups the segments. For each group, a fused wall is generated by averaging the polar parameters: the mean distance $\bar{\rho}$ and the mean normal angle $\bar{\alpha}$ (computed by the circular mean of $sin\alpha$ and $cos\alpha$). The endpoints of the fused wall are then determined by projecting all original endpoints belonging to the cluster onto the line defined by $(\bar{\rho}, \bar{\alpha})$ and taking the minimum and maximum projected distances along the line direction. This yields a single consolidated line segment that accurately represents the underlying wall.
\end{enumerate}

Consequently, this module integrates the raw segment detections and outputs spatially organized and fused wall vector features. Additionally, for semantic detectors such as YOLO, the module further preserves the center coordinates and radius information associated with detected columns.

\subsubsection{Multi-Epoch Fusion}
\label{subsub:multi_epoch_fusion}

The Global Fusion Node integrates the locally fused features into a persistent global map over time.

\begin{enumerate}
    \item Temporal Data Association:
    Incoming local features are matched against existing global features using a Matching Score based on Euclidean distance, angular alignment, and overlap ratio. A \textit{Best Match} logic determines correspondences. We apply \textit{Greedy Algorithm} here.

    \item Lifecycle Management:
    To robustly distinguish between static structures and dynamic noise (e.g. moving pedestrians), a lifecycle mechanism is implemented:
    \begin{itemize}
        \item Initialization: Unmatched local features create new global feature candidates with an initial covariance.
        \item Survival (Hits): If a feature is successfully matched, its \texttt{hit} counter increases.
        \item Pruning (Misses): If a feature is expected to be visible but is not matched, its \texttt{miss} counter increments. Features exceeding the threshold are considered transient noise and removed.
    \end{itemize}

    \item State Estimation and Update:
    For matched features, we employ a recursive filter to update their geometric parameters. Specifically, for wall features, a standard Kalman Filter is applied. The state parameters are defined in polar coordinates: $(\rho, \alpha, d_1, d_2)$. Both the initial state covariance matrix and the measurement matrix are initialized as $diag(\sigma_r^2,\sigma_\alpha^2,\sigma_r^2,\sigma_r^2)$, based on the LiDAR's ranging error $\sigma_r$ and the angular error $\sigma_\alpha$. By updating the state vector and the covariance matrix through the filter, the variance of the wall positions and angles is progressively reduced as more observations are accumulated. In contrast, for column elements detected by YOLO, a simple iterative averaging method is applied.

    \item Manhattan World Optimization:
     Using the inherent structural regularity of indoor rooms, we develop a dedicated Manhattan World Optimizer. This module bridges the semantic information of wall segments to room semantic recognition results via three tightly coupled key steps: dominant direction estimation, Manhattan constraint application, and wall closure. The process is explained in Figure~\ref{fig:manhattan_optimizer}.

    Specifically, the submodule first estimates the dominant directions (principal axes) of the environment. It then applies orthogonal constraints to the global map, snapping walls that are nearly parallel or perpendicular to the dominant directions to perfect orthogonality. Finally, a corner closure logic re-calculates intersections based on the optimized wall lines and merges adjacent endpoints, resulting in a closed, watertight semantic map.

    \begin{figure}[htbp]
        \centering
        \includegraphics[width=\textwidth]{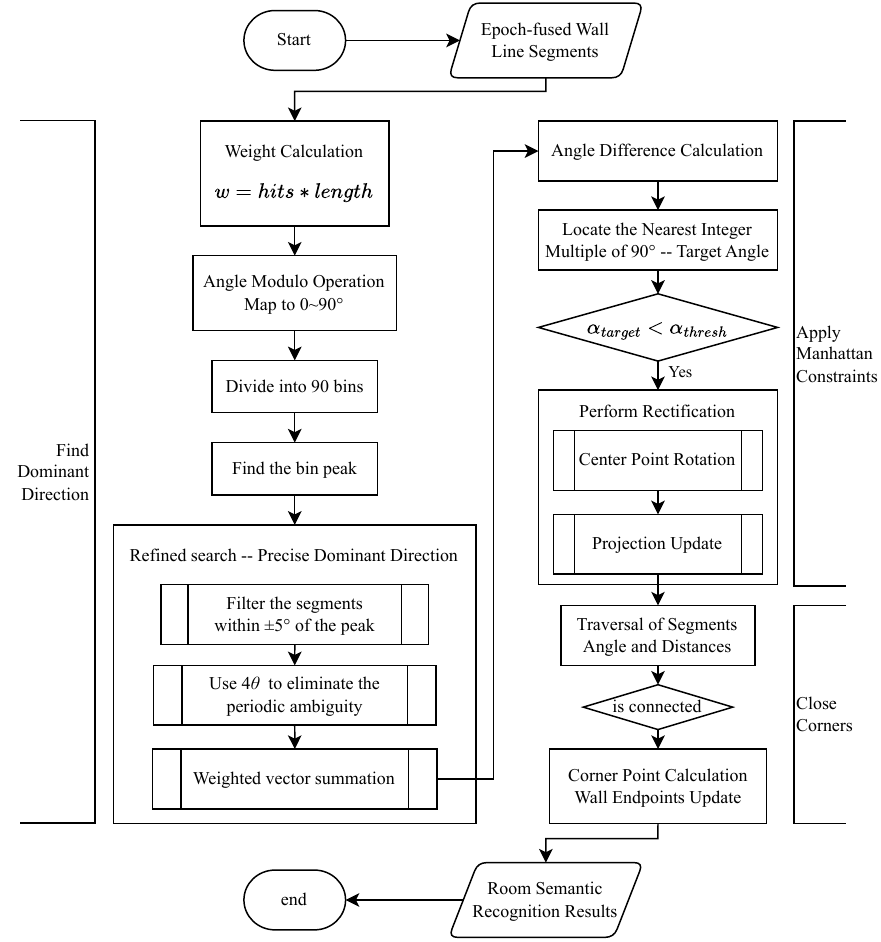}
        \caption[Pipeline of the Manhattan World Optimizer]{\textbf{Pipeline of the Manhattan World Optimizer.} Leveraging the inherent Manhattan structural prior of indoor scenes, this optimizer integrates scattered semantic information of wall segments into a high-precision, topologically closed indoor room floorplan.}
        \label{fig:manhattan_optimizer}
    \end{figure}
\end{enumerate}

Consequently, this module integrates locally fused features into a consistent global map and generates a geometrically accurate global structured map with strong robustness to dynamic noise via temporal data association, lifecycle management, and iterative state estimation. In addition, the unified polar parameterization of wall features not only adapts to the Kalman filter update pipeline of this module but also provides a standardized parameter interface for subsequent global pose optimization and back-end graph-based optimization in dynamic scenes.

\section{Experimental Validation}
\label{sec:experimental-validation}

This section presents a comprehensive experimental validation of the proposed feature extraction and mapping pipeline. To assess the system's practical viability, comparative experiments are conducted across diverse real-world indoor scenarios. The analysis focuses on benchmarking the implemented multi-paradigm strategies in terms of semantic fidelity, geometric accuracy, and computational efficiency. Furthermore, the system's real-time capability is rigorously tested on a standard mobile computing platform to demonstrate its suitability for resource-constrained onboard deployment.

\subsection{Experimental Setup}
\label{sub:experimental_setup}

\subsubsection{Experimental Platform and Scenarios}
\label{subsub:experimental_platform_and_scenarios}

To assess the system's accuracy performance and for the convenience of obtaining experimental results, the results on accuracy were conducted on Lenovo ThinkBook 14 G7. The system is powered by an Intel Core Ultra 5 125H processor and 32 GB LPDDR5x RAM, using integrated graphics without discrete GPU acceleration. This setup is chosen to simulate the computational constraints of standard service robots. The software framework runs on Ubuntu 22.04 LTS with ROS 2 Humble. 

Meanwhile, the timing results were obtained by running the system on a Raspberry Pi 5. The device operates on Raspberry Pi OS (64-bit) with 8 GB of memory and 4 cores and without GUI. The ROS 2 packages were deployed to the device via Docker, and GPU was not utilized. Crucially, to rigorously analyze system latency, the default Quality of Service (QoS) profile for sensor topics was configured to \textit{Best Effor} with a message history depth of 1.

Perception data was acquired using an Ouster OS2 long-range 3D LiDAR sensor, which offers a range precision of $\pm2$ cm.

\begin{figure}[htbp]
    \centering
    \includegraphics[width=0.8\textwidth]{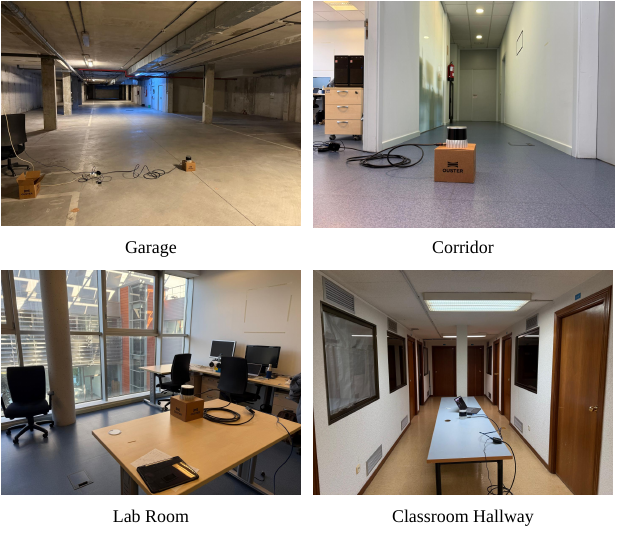}
    \caption[Figures of four scenarios]{\textbf{Figures of four scenarios.} The pipeline is tested within the four scenarios, which range from a few meters to tens of meters.}
    \label{fig:scenario_img}
\end{figure}

To verify the robustness of the proposed method across diverse environments, experimental data was collected in four distinct scenarios: a large-scale underground garage, a long narrow corridor, a cluttered laboratory, and a classroom hallway, as shown in Figure~\ref{fig:scenario_img}. These scenes range from a few meters to tens of meters in scale, with progressively increasing clutter and noise levels, effectively covering the operational domain of indoor mobile robots. Ground truth maps were manually annotated and verified using a high-precision ultrasonic distance sensor (Bosch DUS 20) to ensure metric accuracy.

\subsubsection{Evaluation Metrics}
\label{subsub:evaluation_metrics}

We adopted a multi-dimensional metric system to evaluate both computational efficiency and mapping accuracy.

First, efficiency metrics are conducted in the largest scenario garage, and also include the following items:

\begin{enumerate}
    \item End-to-End Latency: The total time from sensor data acquisition to global map update.
    \item Inference Time: The standalone processing time of the feature extraction algorithm.
    \item Module Latency: Specific consumption of the Local Fusion and Global Fusion nodes to identify bottlenecks.
    \item CPU and Memory Usage: The percentage of CPU and memory consumption during peak operation.
\end{enumerate}

Then accuracy metrics are evaluated across all four scenarios using:

\begin{enumerate}
    \item Length-Weighted Recall: The ratio of the length of correctly detected wall segments to the total length of ground truth walls.
    \item Length-Weighted Precision: The ratio of the length of correctly detected wall segments to the total length of all detected segments.
    \item Geometric Error: Decoupled into Perpendicular Distance Error (deviation of the midpoint) and Angular Error (orientation mismatch). Both are weighted by the length of the wall segments to prioritize dominant structural elements over fragmented noise.
\end{enumerate}

\subsection{Results and Discussion}
\label{sub:results_and_discussion}

\subsubsection{Qualitative Analysis}
\label{subsub:qualitative_analysis}

This section presents a qualitative analysis of the output results of the four methods in the scenarios. To better illustrate the experimental site and detection results, a comprehensive video demonstration comparing these methods is provided at \url{https://vimeo.com/1174443771?share=copy&fl=sv&fe=ci}.

In the structurally simple underground garage scenario, all four evaluated methods successfully achieve effective wall identification and floor plan construction. In particular, the YOLO-based approach demonstrates semantic capability by distinguishing and localizing pillars within the open space.

\begin{figure}
    \centering
    \includegraphics[width=1\linewidth]{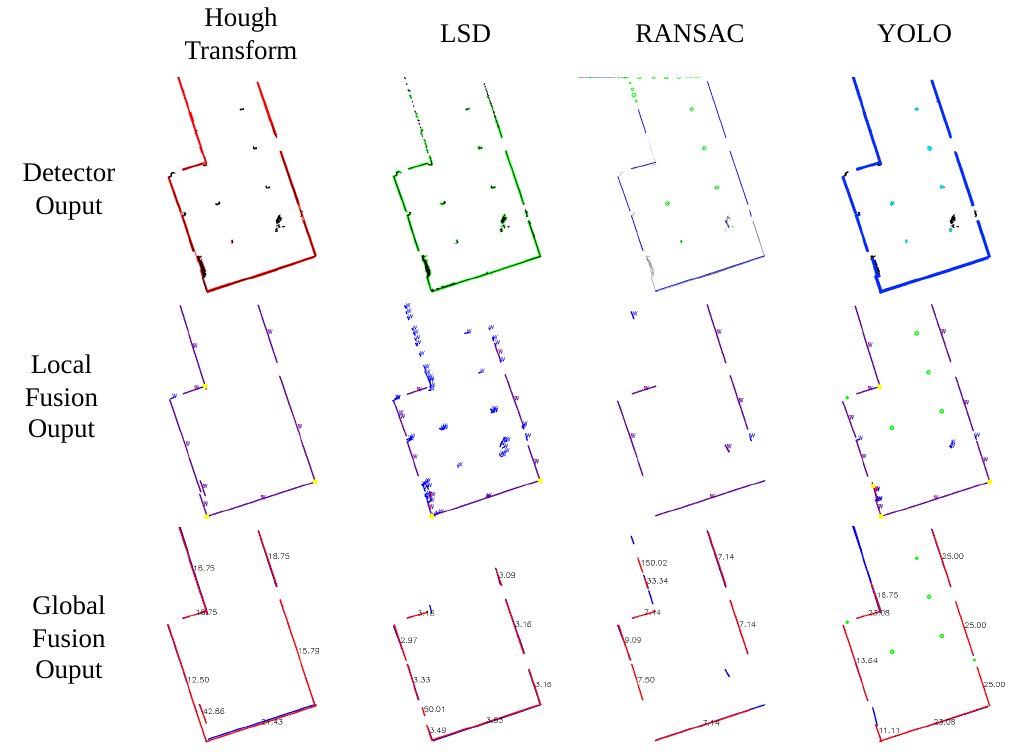}
    \caption[Performance Comparison in Garage Scenario]{\textbf{Performance Comparison of the Four Methods in Garage Scenario across Three Processing Stage.} The figure shows the outputs of three nodes: Detector, Local Fusion, and Global Fusion. For Detector Output, the black 2D point cloud represents the 3D LiDAR BEV map, overlaid with wall recognition results. Notably, for RANSAC, green marking indicate line segments rejected as too short, while for YOLO, cyan markings identify columns. For Local Fusion Output, blue line segments represent the raw input received from the Detector Node, while red segments depict the spatially fused wall structures. For Global Fusion Output, the blue segments correspond to the input from the Local Fusion Node, and red segments illustrate the final temporally fused global map. Numerical annotations on the segments indicate the rank of the observation matrix, reflecting estimation stability.}
    \label{fig:garage_result}
\end{figure}

Figure~\ref{fig:garage_result} presents a comprehensive visualization of the processing pipeline in the Garage scenario. The detector identification images display the BEV point cloud overlaid with the recognition results of each method. These results serve as the input for the Local Fusion node, depicted as blue line segments, while the locally fused outputs are shown in red. Subsequently, these locally fused features act as inputs for the Global Fusion node, again represented in blue, with the final temporally fused global map shown in red. The numerical values annotated on the line segments indicate the rank of the observation matrix, serving as an indicator of observational convergence and estimation stability.

\begin{figure}
    \centering
    \includegraphics[width=0.8\linewidth]{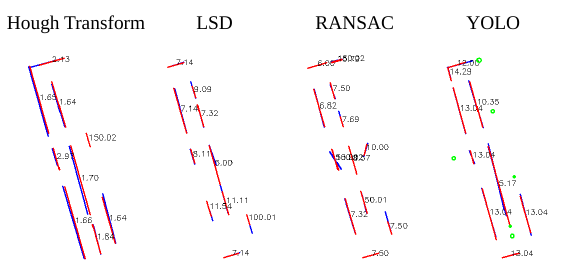}
    \caption[Global Fusion Outputs for the Corridor Scenario.]{\textbf{Global Fusion Outputs for the Corridor Scenario.} The presence of glass surfaces causes specular reflections, resulting in geometric artifacts resembling a parallel corridor.}
    \label{fig:corridor_result}
\end{figure}

The Corridor scenario introduces perceptual challenges due to extensive glass surfaces. These surfaces induce specular reflections in the LiDAR data, creating phantom artifacts that appear as a parallel corridor adjacent to the real structure. Figure~\ref{fig:corridor_result} illustrates the final Global Fusion outputs for all four methods, highlighting their respective abilities to handle such environmental ambiguities.

\begin{figure}
    \centering
    \includegraphics[width=\linewidth]{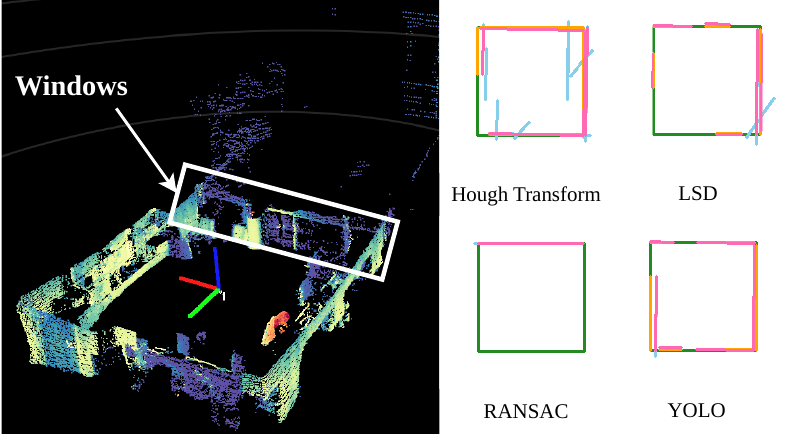}
    \caption[Raw Point Cloud and Global Fusion Outputs for the Laboratory Scenario.]{\textbf{Raw Point Cloud and Global Fusion Outputs for the Laboratory Scenario.} The left panel displays the raw 3D point cloud. The right panels display a 2$\times$2 comparison of the four methods against the ground truth. Legend: Green lines represent the Ground Truth; Pink lines indicate successfully matched detections (True Positives); Blue lines represent unmatched detections (False Positives); Orange lines correspond to the matched segments of the Ground Truth.}
    \label{fig:cslab_results}
\end{figure}

The Laboratory environment presents a highly cluttered setting containing furniture, equipment, and shelves. Furthermore, a full-length glass wall results in a sparse point cloud density along that boundary, reducing the detection quality. Figure~\ref{fig:cslab_results} compares the raw point cloud with the mapping results against the ground truth, applying a range filter to exclude points outside the glass wall. Under these conditions, the Hough Transform and the LSD generate significant noise caused by clutter edges. RANSAC fails to converge on valid models due to the low inlier ratio. Conversely, the YOLO approach effectively filters non-structural elements and provides the most accurate reconstruction.

\begin{figure}
    \centering
    \includegraphics[width=\linewidth]{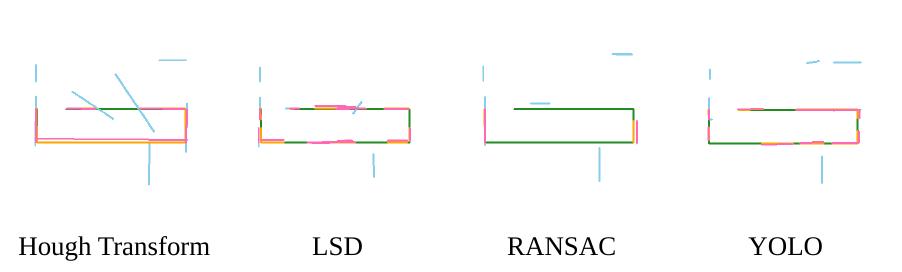}
    \caption[Global Fusion Outputs for the Classroom Hallway Scenario.]{\textbf{Global Fusion Outputs for the Classroom Hallway Scenario.} The comparison of outputs against the ground truth is shown as well. While geometric methods capture fine details, the deep learning approach exhibits lower recall in this specific setting.}
    \label{fig:classroom_hallway_result}
\end{figure}

The Classroom Hallway exhibits a similarly complex environment characterized by glass partitions that reveal clutter in adjacent rooms. In this specific scenario, the Hough Transform and LSD unexpectedly outperform the data-driven approach although still contain noisy lines. RANSAC remains ineffective due to the complex outlier distribution. The YOLO model succeeds in noise elimination, but struggles to recall wall features in this scenario, a phenomenon attributed to feature resolution mismatches that will be analyzed in detail in Section~\ref{subsub:discussion_on_system_robustness_and_generalization}. The comparison between the generated maps and the ground truth is shown in Figure~\ref{fig:classroom_hallway_result}.

\subsubsection{Quantitative Performance Comparison}
\label{subsub:quantitative_performance_comparison}

This section provides a systematic evaluation of the computational efficiency and detection accuracy of the proposed system.

Firstly, the analysis of computational efficiency and resource consumption is conducted in the Garage scenario to ensure valid detections across all methods, providing a fair baseline for resource benchmarking. The hardware equipment and software configuration used in the experiment are Raspberry Pi 5, as described in Section~\ref{subsub:experimental_platform_and_scenarios}.

The test results of four evaluated methods are presented. For the YOLO model, we tested multiple compilation and inference frameworks, including OpenVINO, NCNN, and TensorFlow Lite (TFLite). The model quantization schemes were tested with FP32 and INT8, while image sizes of 1024$\times$1024 and 640$\times$640 were examined. After comprehensively balancing computational efficiency and accuracy performance, we select the results of the FP32-precision YOLO model based on the OpenVINO framework for presentation herein, with test data for both image resolutions included. 

\begin{figure}
    \centering
    \includegraphics[width=0.8\linewidth]{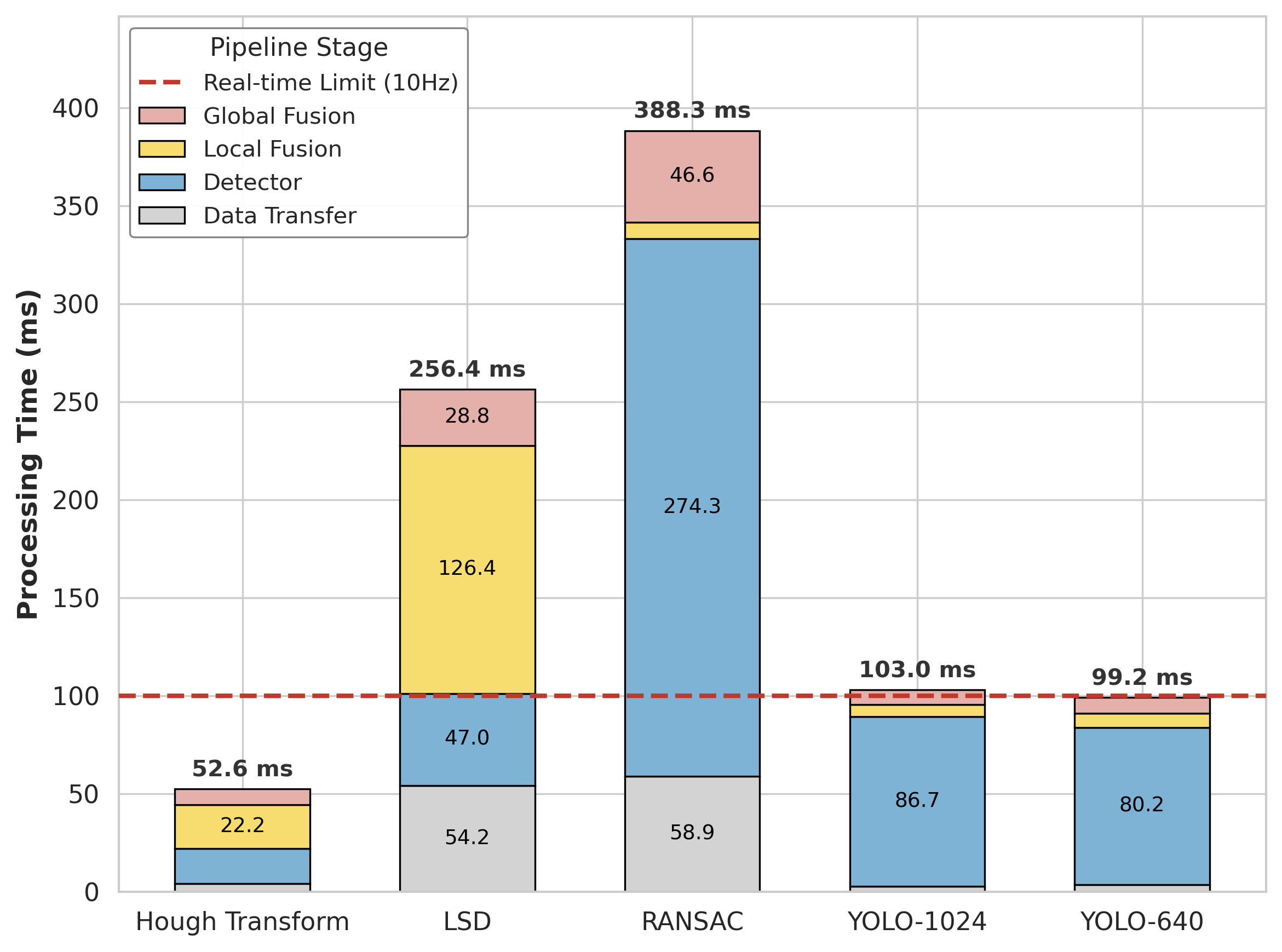}
    \caption[Latency Breakdown Per Frame in the Garage Scenario.]{\textbf{Latency Breakdown Per Frame in the Garage Scenario.} The stacked bars illustrate the time consumption of the Detector, Local Fusion, Global Fusion, and Data Transfer components. The corresponding methods include the Hough Transform, LSD, RANSAC, and the YOLO-OBB method with image size of 1024 and 640. The red dashed line represents the 10 Hz real-time threshold (100 ms).}
    \label{fig:Fig1_Latency_Breakdown}
\end{figure}

Figure~\ref{fig:Fig1_Latency_Breakdown} illustrates the breakdown of the processing latency per frame.
\begin{enumerate}
    \item Real-time Capability: The Hough Transform, and YOLO with image size of 640 all operate well below the 100 ms real-time threshold. For the YOLO method with an image size of 1024 pixels, its end-to-end processing time is just over 3 milliseconds.
    \item Latency Distribution: A distinct structural difference is observed. Traditional geometric methods rely on lightweight detection but incur higher overheads in data transfer and fusion due to the large volume of fragmented primitives. In contrast, the YOLO-based approach shifts the computational load to the Detector stage (over 80\%). However, because the network outputs high-quality, semantically filtered object bounds, the subsequent Local and Global Fusion stages are extremely efficient (consuming negligible time).
    \item Justification for Dimensionality Reduction: In fact, 3D point cloud-based plane detection methods (such as 3D RANSAC, Region Growing, etc.) operate on a different data processing scale compared with these methods, making a direct comparison unfair. Quantitative experimental results presented here demonstrate that the inference time of 3D point cloud-based plane detection methods is more than 10 times that of YOLO.
\end{enumerate}

\begin{figure}[htbp]
    \centering
    \includegraphics[width=\textwidth]{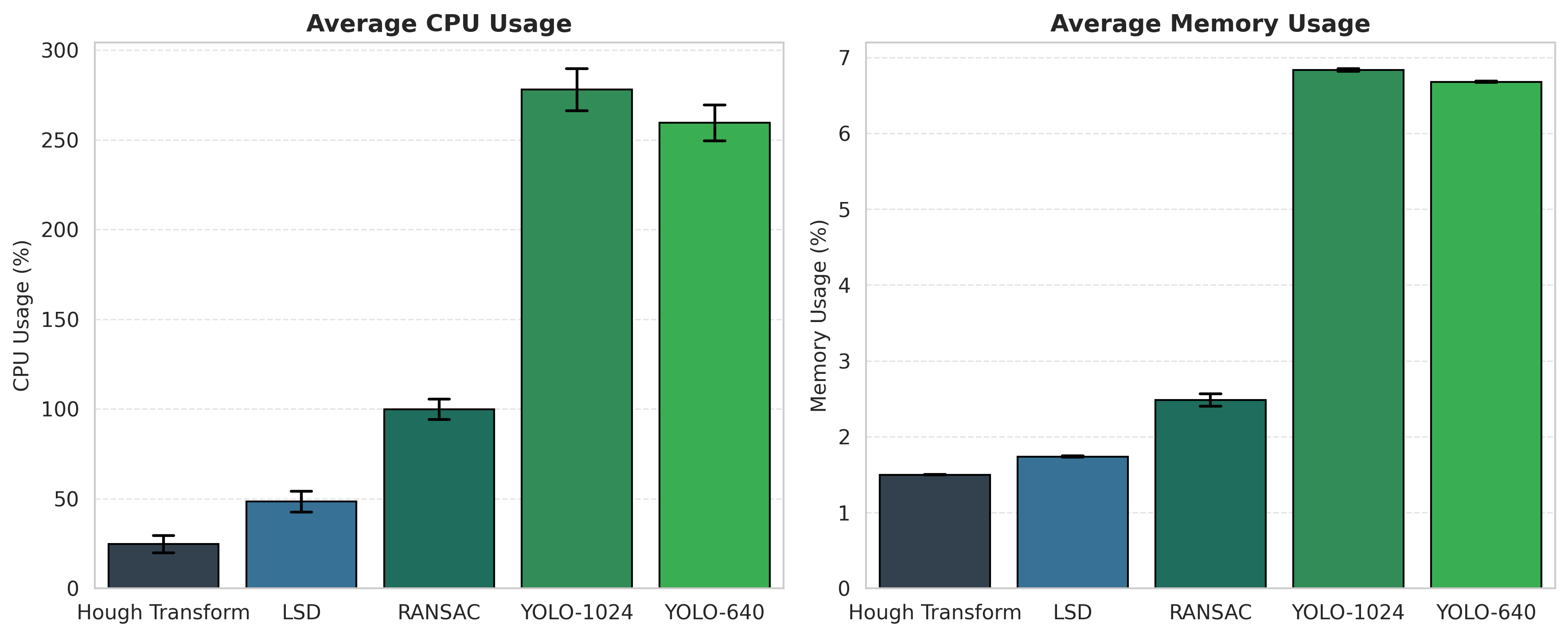}
    \caption[Average Resource Consumption comparison.]{\textbf{Average resource consumption comparison on the mobile computing platform.} (Left) Average CPU usage percentage, where RANSAC exhibits excessive load due to iterative calculations. (Right) System memory usage percentage. Although YOLO-OBB requires higher memory allocation for the neural network, they remain below 7\% of the total available RAM (8GB), proving its feasibility for onboard deployment.}
    \label{fig:Fig2_Resource_Usage}
\end{figure}

In terms of resource utilization, as shown in Figure~\ref{fig:Fig2_Resource_Usage}, YOLO exhibits relatively high CPU utilization, reaching approximately 270\%, which is 2.7 times that of the second-ranked RANSAC. Regarding the memory footprint, while deep learning models inherently require more memory to store weights and activation maps, the YOLO-OBB systems consume only 7\% of the available system memory (approximately 570 MB). This is well within the acceptable limits of modern onboard computers, proving that the neural network-based solution is sufficiently lightweight for embedded deployment.

\begin{figure}[htbp]
    \centering
    \includegraphics[width=\textwidth]{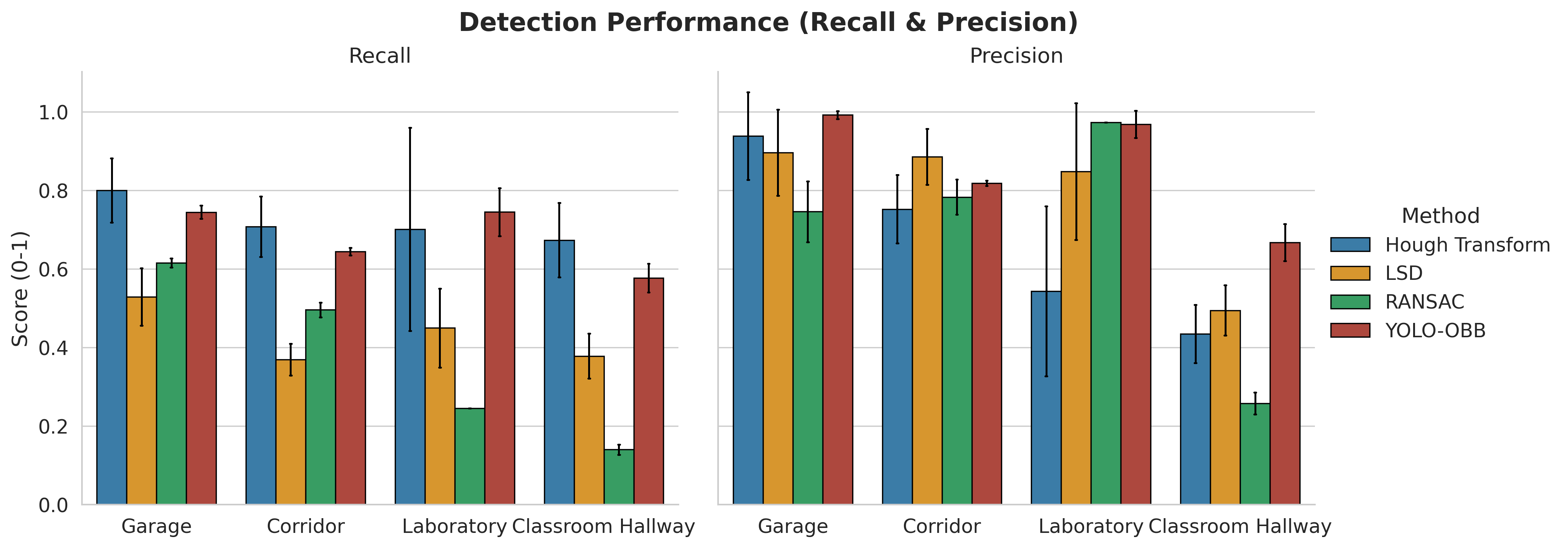} 
    \caption[Detection Performance Evaluation across Four Scenarios.]{\textbf{Detection Performance Evaluation across Four Scenarios.} (Left) Length-Weighted Recall scores. (Right) Length-Weighted Precision scores. Error bars indicate the standard deviation. YOLO-OBB demonstrates superior robustness, particularly in the cluttered Laboratory scenario, while maintaining high precision across all environments.}
    \label{fig:Fig3_Detection_Performance}
\end{figure}

Secondly, detection performance is comprehensively analyzed using Recall, Precision, and F1-Score metrics across four diverse scenarios using the calculation platform of ThinkBook, as shown in Figure~\ref{fig:Fig3_Detection_Performance} and Table~\ref{tab:f1_score_summary}. The most significant advantage of YOLO-OBB is demonstrated in the Laboratory scenario, which features intense environmental clutter. In this challenging setting, YOLO achieves the highest Recall and Precision, culminating in a superior F1-Score of $0.84 \pm 0.05$, significantly outperforming geometric methods. In contrast, RANSAC exhibits negligible Recall and a poor F1-Score of just 0.39, effectively failing to identify structural elements due to the low inlier ratio caused by clutter, as qualitatively observed in Figure~\ref{fig:cslab_results}. Furthermore, across most scenarios, YOLO maintains the highest Precision. This indicates that the system generates minimal false positives, effectively filtering out noise and transient obstacles. This robustness is further evidenced in the Garage and Corridor scenarios; while YOLO-OBB achieves F1-Scores ($0.85$ and $0.72$, respectively) comparable to the Hough Transform, it exhibits a significantly lower standard deviation ($0.01$), demonstrating exceptional stability. Regarding the Classroom Hallway scenario, while the Hough Transform achieves a slightly higher Recall, it suffers from poor Precision (approximately 0.4). This severe noise contamination drags its F1-Score down to 0.53. In contrast, YOLO maintains high Precision while achieving a competitive Recall (approx. 0.6), ultimately securing the highest F1-Score ($0.62 \pm 0.04$) in this environment. Although a minor domain gap with respect to thin wall features persists, the high precision and overall superior F1-Scores ensure that the detected structures are trustworthy, avoiding the map pollution observed in geometric methods.

\begin{table}[htbp]
    \centering
    \caption{\textbf{Quantitative Detection Performance (F1-Score) across Scenarios.} The table presents the average and standrad deviation of F1-Score to evaluate the overall detection quality.}
    \label{tab:f1_score_summary}
    \small
    \renewcommand{\arraystretch}{1.3}
    \begin{tabular}{l cccc}
        \toprule
        \multirow{2}{*}{\textbf{Method}} & \multicolumn{4}{c}{\textbf{Scenario}} \\
        \cmidrule(lr){2-5}
        & \textbf{Garage} & \textbf{Corridor} & \textbf{Laboratory} & \textbf{Classroom Hallway} \\
        \midrule
        Hough Transform & \textbf{0.86 $\pm$ 0.09} & \textbf{0.73 $\pm$ 0.08} & 0.61 $\pm$ 0.23 & 0.53 $\pm$ 0.08 \\
        LSD             & 0.66 $\pm$ 0.07 & 0.52 $\pm$ 0.05 & 0.58 $\pm$ 0.12 & 0.43 $\pm$ 0.05 \\
        RANSAC          & 0.67 $\pm$ 0.03 & 0.61 $\pm$ 0.02 & 0.39 $\pm$ 0.00 & 0.18 $\pm$ 0.02 \\
        YOLO-OBB        & \textbf{0.85 $\pm$ 0.01} & \textbf{0.72 $\pm$ 0.01} & \textbf{0.84 $\pm$ 0.05} & \textbf{0.62 $\pm$ 0.04} \\
        \bottomrule
    \end{tabular}
\end{table}

Finally, the geometric fidelity of the reconstructed maps is evaluated in Table~\ref{tab:geometric_error}. The results highlight a trade-off between peak accuracy and stability. Hough Transform, LSD and RANSAC achieve high accuracy in clean environments, such as the Garage, but degrade significantly in complex scenes like the Classroom Hallway, where distance errors of LSD exceed 15 cm. In contrast, the YOLO method, though the lowest errors are not achieved in all scenarios, shows remarkable consistency. Regardless of the complexity of the environment, its Distance Error remains stable around 6 cm, and the Angle Error stays around 1.0 degrees. It should be noted that the RANSAC method is considered as an algorithm failure in the Laboratory and Classroom Hallway scenarios. This is attributed to the fact that the method could only provide 1 to 2 correctly matched line segments, effectively indicating a failure of the algorithm in these contexts, as shown in Figure~\ref{fig:cslab_results} and Figure~\ref{fig:classroom_hallway_result}.

\begin{table}[htbp]
    \centering
    \caption{\textbf{Geometric Accuracy Assessment of the Reconstructed Maps.} The table compares the Perpendicular Distance Error (cm) and Angular Error (degrees) across four scenarios. Note that ``-'' indicates an algorithm failure where insufficient line segments were matched (specifically RANSAC in complex environments).}
    \label{tab:geometric_error}
    \begin{threeparttable}
        \resizebox{\textwidth}{!}{
            \begin{tabular}{lcccccccc}
                \toprule
                \multirow{2}{*}{\textbf{Scenario}} & \multicolumn{4}{c}{\textbf{Distance Error (cm)} $\downarrow$} & \multicolumn{4}{c}{\textbf{Angle Error (deg)} $\downarrow$} \\
                \cmidrule(lr){2-5} \cmidrule(lr){6-9}
                 & Hough & LSD & RANSAC & \textbf{YOLO} & Hough & LSD & RANSAC & \textbf{YOLO} \\
                \midrule
                Garage            & 6.47 & 4.65 & \textbf{3.47} & 6.03 & 0.25 & 0.23 & 0.14 & \textbf{0.09} \\
                Corridor          & 10.36 & 7.03 & \textbf{5.34} & 6.15 & 0.41 & 0.24 & \textbf{0.10} & 0.35 \\
                Laboratory        & 13.16 & \textbf{6.80} & - & 9.38 & 1.52 & 1.42 & - & \textbf{1.09} \\
                Classroom Hallway & 10.17 & 15.38 & - & \textbf{5.73} & 0.95 & \textbf{0.73} & - & 1.12 \\
                \midrule
                \textbf{Average}  & 10.02 & 8.46 & N/A & \textbf{6.83} & 0.77 & 0.66 & N/A & \textbf{0.66} \\
                \bottomrule
            \end{tabular}
        }
    \end{threeparttable}
\end{table}

\subsubsection{Discussion on System Robustness and Generalization}
\label{subsub:discussion_on_system_robustness_and_generalization}

Based on the qualitative and quantitative analysis presented above, a distinct performance dichotomy emerges between the geometric and neutral network paradigms. As summarized in Table~\ref{tab:qualitative_summary_compact}, traditional geometric methods (LSD, Hough, and RANSAC) demonstrate competence in large-scale, sparse environments like the Garage, where they provide high geometric detail. However, their lack of semantic understanding results in low robustness in complex, cluttered scenarios (e.g., the Laboratory). In such settings, these methods are prone to catastrophic failure caused by "feature blockage," where the edges of furniture are misidentified as structural walls.

In contrast, the proposed YOLOv8n-obb model exhibits high robustness and stability across all tested domains. Using learned semantic representations, it effectively filters out transient clutter and retains only the relevant structural topology (walls and columns). To further elucidate the operational boundaries and deployment constraints of these algorithms, we analyze three critical technical phenomena observed during the experiments:

\begin{table}[htbp]
    \centering
    \caption{\textbf{Qualitative Summary of Method Performance}}
    \label{tab:qualitative_summary_compact}
    \small
    \renewcommand{\arraystretch}{1.3}
    \begin{tabularx}{\linewidth}{l X X c}
        \toprule
        \textbf{Method} & \textbf{Large Simple Scenario} & \textbf{Complex Scenario} & \textbf{Robustness} \\
        \midrule
        Hough & Efficient; lacks semantic awareness. & High false positives due to clutter. & Medium \\
        LSD & High geometric detail; fast. & Severe over-segmentation; computational bottleneck. & Medium \\
        RANSAC & Good modeling; high latency. & Convergence failure due to low inlier ratios. & Low \\
        \textbf{YOLO} & \textbf{Accurate classification} of structural elements. & \textbf{Excellent clutter rejection}; high stability. & \textbf{High} \\
        \bottomrule
    \end{tabularx}
\end{table}

In addition, we observed some interesting phenomena that is worth being discussed.

\begin{enumerate}
    \item Analysis of Over-Segmentation in LSD: While LSD theoretically offers high precision with high speed, experimental results reveal a critical limitation in complex environments. As observed in the cluttered laboratory scenario, LSD tends to fracture continuous walls into numerous short, disjointed line segments. As shown in Figure~\ref{fig:oversegmentforlsd}, the over-segmentation increases the complexity of post-process including backend spatial clustering (DBSCAN) and matching algorithms. It causes a significant drop in processing speed, which indicates that the method is not suitable for complicated scenarios.
    \begin{figure}
        \centering
        \includegraphics[width=0.7\linewidth]{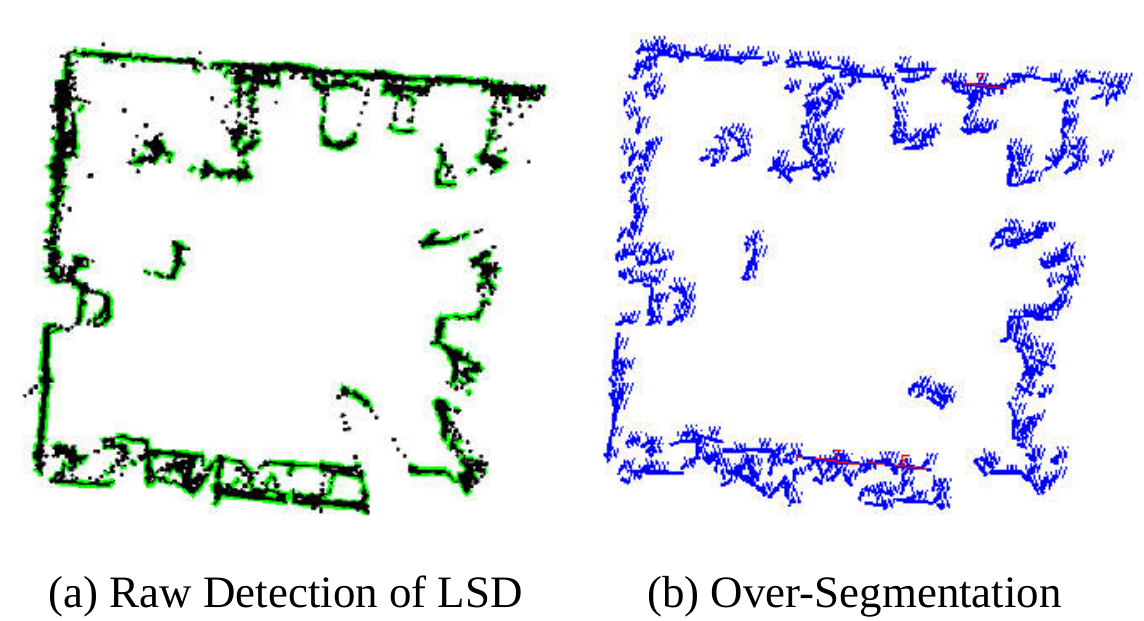}
        \caption[Over-segmentation of the LSD Method.]{\textbf{Over-segmentation of the LSD Method.} (Left) It shows the original point cloud BEV map directly processed by the LSD method. (Right) It presents the corresponding extracted line segments, which are extremely numerous.}
        \label{fig:oversegmentforlsd}
    \end{figure}
    \item Impact of Scene Size on Feature Resolution: The method leverages the noise characteristics of LiDAR, where vertical walls typically appear as dense, thick rectangular clusters in BEV projections. For smaller-scale scenarios, the wall features appeared to be much thinner due to the shorter scanning distance and lower absolute noise. These lines differ from the walls in the synthetic training set for the YOLO detector. Consequently, the model initially struggled to recall these thin features, as shown in the Figure~\ref{fig:thinwall}, while geometric methods work better in situations. Therefore, it can be argued that with the supplementation of the training set and the improvement of synthetic data, the YOLO method has greater potential.
    \begin{figure}
        \centering
        \includegraphics[width=0.5\linewidth]{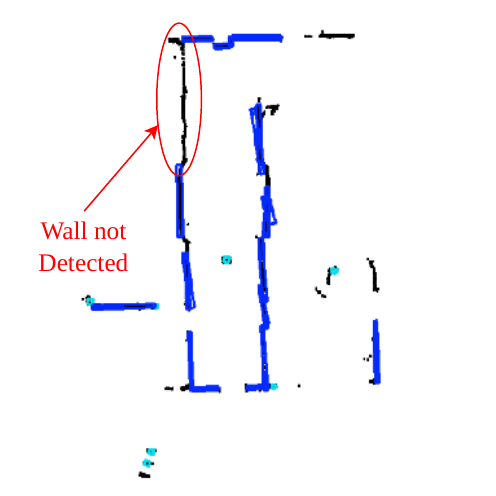}
        \caption[False Negative Detection due to Feature Resolution Mismatch.]{\textbf{False Negative Detection due to Feature Resolution Mismatch.} The figure illustrates a specific failure case in Classroom Hallway, a small-scale scenario. The blue lines represent the model's predictions overlaid on the raw point cloud BEV. The red circle highlights a distinct, thin wall segment that the model failed to detect (False Negative). This omission is attributed to the domain gap between the thick, noisy walls in the synthetic training data and the sharp, thin features in this real-world scan.}
        \label{fig:thinwall}
    \end{figure}
\end{enumerate}
\section{Conclusion}
\label{sec:conclusion}

This study addresses the critical challenge of constructing semantic structural maps online using resource-constrained onboard robotic platforms. By leveraging the projection of 3D LiDAR data into 2D BEV images, we proposed a lightweight and modular perception framework. To identify the optimal processing strategy for this pipeline, we systematically evaluated a spectrum of lightweight algorithms, directly comparing classical geometric methods against a data-driven neural network (YOLO-OBB) approach.

Comprehensive experimental validations across diverse environmental complexities revealed a significant performance trade-off. While classical geometric approaches offer fast initial detection, they are highly susceptible to environmental noise and clutter, leading to steep declines in precision and overall detection quality (as supported by the performance metrics in Figure~\ref{fig:Fig3_Detection_Performance} and F1-Scores in Table~\ref{tab:f1_score_summary}). Furthermore, this susceptibility generates fragmented and unstable outputs that paradoxically increase the computational burden on downstream fusion modules. As evidenced by the latency breakdown (Figure~\ref{fig:Fig1_Latency_Breakdown}), this backend congestion severely compromises real-time operation. Conversely, the proposed deep learning-based pipeline effectively filters non-structural noise at the detection stage through semantic understanding. By outputting high-quality, stable bounding boxes, it drastically reduces the backend computational overhead, maintaining robust real-time performance even on low-power edge devices without GPU acceleration.

Ultimately, robustness evaluations demonstrate that geometric methods lack stability in complex real-world scenarios, exhibiting precipitous drops in precision as detailed in Figure~\ref{fig:Fig3_Detection_Performance}. The YOLO-OBB-based approach, however, maintains consistent geometric fidelity and superior noise immunity across all tested environments. We conclude that this data-driven strategy achieves an optimal balance between computational efficiency and perceptual robustness, proving to be a highly effective and deployable solution for autonomous indoor navigation on embedded robotic systems.

\section*{Acknowledgments}
\label{sec:acknowledgments}

Funded by the European Union. The views and opinions expressed are, however, those of the author(s) only and do not necessarily reflect those of the European Union or the European Commission. Neither the European Union nor the granting authority can be held responsible for them.
This work received funding from the European Union’s Horizon Europe programme under grant agreement ID: 101136056 (SHEREC) under the Horizon Europe Program HORIZON-CL4-2023-HUMAN-01 CNECT.

\section*{Data and Code Availability}
The datasets and source code generated and/or analyzed during the current study are available in the GitLab repository, \url{https://gitlab.com/cvar-upm/release/lidar-bev-structural-detector}.

\begin{adjustwidth}{-\extralength}{0cm}



\bibliographystyle{unsrt}
\bibliography{70_reference}

\end{adjustwidth}
\end{document}